\documentclass{article}

\usepackage[final]{timeseries_workshop}

\usepackage[utf8]{inputenc} 
\usepackage[T1]{fontenc}    
\usepackage{hyperref}       
\usepackage{url}            
\usepackage{booktabs}       
\usepackage{amsfonts}       
\usepackage{nicefrac}       
\usepackage{microtype}      
\usepackage{xcolor}         

\title{Efficiently Generating Correlated Sample Paths from Multi-step Time Series Foundation Models}

\author{%
  Ethan Baron\thanks{Work completed while at Amazon. Correspondence to \texttt{eb4727@nyu.edu}.} \\
  New York University
  \And
  Boris Oreshkin \\
  Amazon
  \And
  Ruijun Ma \\
  Amazon
  \And
  Hanyu Zhang \\
  Amazon
  \And
  Kari Torkkola \\
  Amazon
  \And
  Michael W. Mahoney \\
  Amazon
  \And
  Andrew Gordon Wilson \\
  New York University \& Amazon
  \And
  Tatiana Konstantinova \\
  Amazon
}

\usepackage{graphicx}

\usepackage{titlesec}
\titlespacing{\section}{0pt}{\parskip}{0pt}

\usepackage{xurl}

\usepackage{multirow}

\begin{document}

\maketitle

\begin{abstract}
Many time series applications require access to multi-step forecast trajectories in the form of sample paths. 
Recently, time series foundation models have leveraged multi-step lookahead predictions to improve the quality and efficiency of multi-step forecasts. 
However, these models only predict independent marginal distributions for each time step, rather than a full joint predictive distribution. 
To generate forecast sample paths with realistic correlation structures, one typically resorts to autoregressive sampling, which can be extremely expensive. 
In this paper, we present a copula-based approach to efficiently generate accurate, correlated sample paths from existing multi-step time series foundation models in one forward pass. 
Our copula-based approach generates correlated sample paths orders of magnitude faster than autoregressive sampling, and it yields improved sample path quality by mitigating the snowballing error phenomenon.
\end{abstract}

\section{Introduction}

In recent years, time series foundation models (TSFMs) have received significant attention, in some cases outperforming statistical and deep learning methods on some public time series forecasting benchmarks.
One common characteristic of top-performing TSFMs, including TimesFM \citep{timesfm}, Chronos-Bolt \citep{chronos_bolt}, and TiRex \citep{tirex}, is their ability to make multi-step predictions in a single forward pass. 
That is, given an input time series $x_{1:T}$, these \emph{multi-step models}  produce probabilistic forecasts $p_\theta(x_{T+i} \mid x_{1:T})$ simultaneously for multiple future horizons $i = 1, 2, \ldots, H$. This feature has allowed recent TSFMs to make improved predictions over longer horizons, and it has increased the efficiency of both training and inference.

Multi-step TSFMs produce \emph{marginal} predictive distributions $p_\theta(x_{T+i} \mid x_{1:T})$, rather than a full \emph{joint} predictive distribution $p(x_{T+1 \, : \, T+H} \mid x_{1:T})$. 
This is an important limitation because in many practical scenarios, having access to a full joint distribution is crucial \citep{joint}. 
For instance, a joint predictive distribution allows practitioners to estimate conditional probabilities, such as $p(x_{T+4} \mid x_{T+2} > c)$, and to estimate statistics over arbitrary aggregation windows, such as the 90th percentile estimate $q_{90}(x_{T+5} + x_{T+6} + x_{T+7})$.
Joint distributions can also be represented implicitly via \emph{sample paths} ---
possible forecast trajectories $x_{T+i \, : \, T+K}$ drawn from the predictive distribution. These sample paths are used as inputs in diverse downstream applications such as supply chain management \citep{sample_paths}, reinforcement learning \citep{rl}, learning dynamic systems \citep{dynamic}, and energy systems \citep{power}.

There are several possible approaches to extract a joint predictive distribution from multi-step TSFMs.
A \emph{naive} approach is to use an independence assumption and set ${p(x_{T+1 \, : \, T+H} \mid x_{1:T}) = \prod_{i=1}^{H} p_\theta(x_{T+i} \mid x_{1:T})}$. 
This approach can be implemented efficiently as it requires only single forward pass of a TSFM. However, the independence assumption is problematic in time series forecasting, where $x_{T+i}$ typically has significant correlation with previous times. For example, if $x_{T+1}$ is relatively high, we would  expect $x_{T+2}$ to also be relatively high. Since the naive approach does not account for these correlations, it results in highly unrealistic sample paths. 

Alternatively, one could use an \emph{autoregressive} sampling strategy by repeatedly sampling one step from the marginal predictive distribution: first sample $\hat{x}_{T+1} \sim p_\theta(x_{T+1} | x_{1:T})$, then ${\hat{x}_{T+2} \sim p_\theta(x_{T+2} | x_{1:T}, \hat{x}_{T+1})}$, and so on. This autoregressive approach allows one to generate realistic sample paths with appropriate correlation structures, but is prohibitively expensive at scale: to generate $N$ sample paths for a horizon length $H$ we must perform $\mathcal{O} (N \cdot H)$ forward passes of the model. It remains unclear how to efficiently produce realistic sample path forecasts with TSFMs.

In this paper, we present a method to generate realistic correlated sample paths from multi-step time series models in one forward pass. 
Our method is based on a \emph{copula} decomposition of the joint predictive distribution, 
allowing us to leverage the accurate marginal predictive distributions of TSFMs and focus our efforts on learning a realistic correlation structure.

\begin{figure}[t!]
    \centering
    \includegraphics[width=1\linewidth]{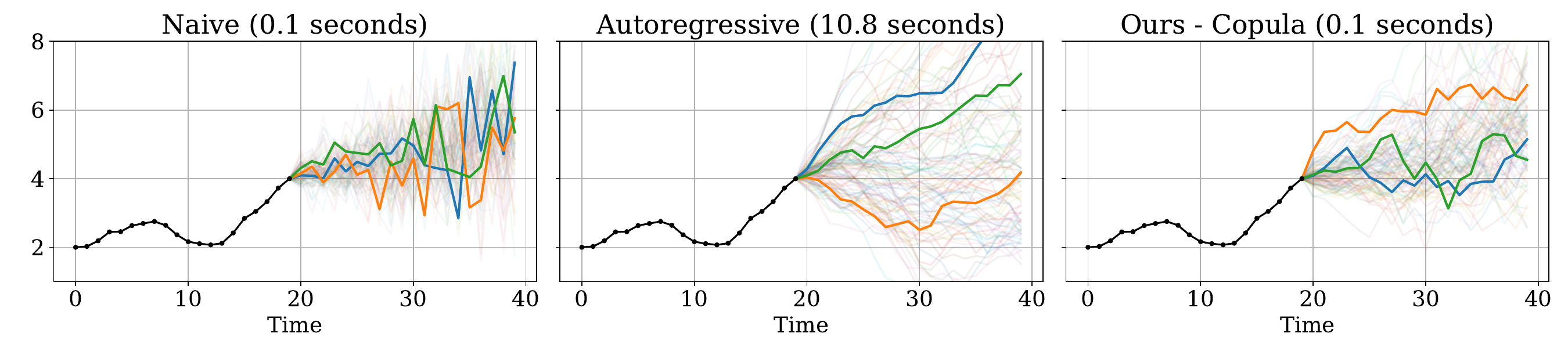}
    \caption{Mechanisms for generating sample paths from multi-step time series models. 100 sample paths are shown for each method, with three randomly-selected sample paths highlighted. The naive approach generates sample paths in $\mathcal{O}(1)$ time but produces jagged trajectories with unrealistic correlation structures. Autoregressive sampling produces smooth correlated trajectories, but takes $\mathcal{O}(N \cdot H)$ forward passes, which can be prohibitive, and can result in snowballing errors. Our copula-based approach provides realistic correlated sample paths in $\mathcal{O}(1)$ time.}
    \label{sample_paths}
    \vspace{-0.5cm}
\end{figure}

The sample paths generated with our method are competitive in quality to those generated autoregressively, but they are obtained significantly more cheaply (see Figure \ref{sample_paths}). 
In fact, we find that the copula-based sample paths are often \emph{more} accurate than autoregressive sampling, achieved by mitigating the \emph{snowballing error phenomenon} \citep{pitfalls}.
Our copula-based approach allows time series practitioners to capitalize on advances in multi-step TSFMs for a wider variety of important use cases, greatly increasing the impact of TSFMs in practice.

\section{Decomposing joint predictive distributions with copulas}
The core idea of our approach leverages the fact that any joint distribution can be factored into a set of marginal distributions and a \emph{copula} function controlling the correlation structure. Formally, consider a collection of random variables $X_1$, \ldots, $X_H$ with marginal CDFs $F_1, \ldots, F_H$, such that ${F_i(x) = p(X_i \leq x)}$. Then, by Sklar's Theorem \citep{sklar}, there exists a copula ${C: [0,1]^H \mapsto [0,1]}$ such that $p(X_1 \leq x_1, \ldots, X_H \leq x_H) = C(F_1(x_1), \ldots, F_d(x_H))$. Intuitively, the copula allows us to decompose the joint distribution of the random variables, modeling their correlation structure in \emph{quantile space} while separately modeling their marginal distributions via $F_i$.

This copula decomposition is particularly compelling for multi-step time series forecasting. We already have access to high-quality marginal distributions $F_i$ obtained from multi-step TSFMs, so all that remains is to model the correlation structure via a copula $C$. Moreover, time series have somewhat predictable correlation structures, such as autocorrelation, which we leverage to obtain a simple yet effective parameterization for the copula.
Several prior works \citep{GMQ-Forecaster, TACTiS, TACTiS-2} use copulas for effective multivariate time series forecasting, whereas our focus is specifically on multi-step forecasting for univariate time series with TSFMs.

\section{A copula-based approach for sample path generation}

We propose a copula-based approach to efficiently generate correlated sample paths from multi-step TSFMs. Specifically, we use a TSFM to construct high-quality marginal predictive distributions for each time step, and augment these marginals with a copula to impose a realistic correlation structure. To generate sample paths with this approach, we first sample vectors of correlated quantiles from the copula and then map these quantiles to the time series space with the marginal predictive distributions.

For the correlation structure, we use a Gaussian copula. Formally, we have ${C(F_1(x_1), \ldots, F_d(x_H)) = \Phi_\Sigma \left(\Phi^{-1}(F_1(x_1)), \, \ldots, \, \Phi^{-1}(F_H(x_H)) \right)}$ where $\Phi^{-1}$ is the inverse CDF of the standard normal distribution and $\Phi_\Sigma$ is the joint CDF of the multivariate normal distribution with mean $\mathbf{0}$ and covariance matrix $\Sigma$. In this work, we parameterize $\Sigma$ as a symmetric Toeplitz matrix with a simple AR(1) correlation structure, such that $\Sigma_{ij} = \rho^{|i-j|}$. This parameterization reflects the autocorrelation structure common in time series. We obtain the parameter $\rho$ by computing the empirical autocorrelation coefficient for each series, $\textrm{Corr}(x_{1:T-1}, x_{2:T})$.
In Appendix \ref{app:pcm}, we consider instead learning the parameterization of $\Sigma$ using a pre-trained lightweight neural network. We find that pre-trained neural networks can predict copula parameterizations that further improve the qualify of sample paths, opening a promising direction for further research.

Note that the TSFMs we consider in this paper do not provide full marginal predictive distributions $F_i$, instead summarizing their marginal predictions to a pre-defined set of quantile knots such as $Q = \{0.1, 0.2, \ldots, 0.9\}$. To convert these predictions to full marginal distributions, we adopt the incremental quantile function approach from \cite{iqf}. Specifically, we fit a full marginal CDF by linearly interpolating between the predicted quantile knots and using exponential decay extrapolation beyond the minimum/maximum knots. We set the decay towards 0 on the left as we exclusively consider non-negative time series in this paper, but this can be adapted for other settings. See Appendix \ref{empirical_cdf} for an illustrative example.

\section{Evaluating multi-step predictions via sample paths}

To evaluate joint predictive distributions, we must capture both the accuracy of the marginal predictive distributions and the realism of the correlation structure across time steps. As is standard, we use the continuous ranked probability score (CRPS) to evaluate the marginal distributions. The CRPS for a joint predictive distribution $p$ and observed vector $\mathbf{x} \in \mathbb{R}^H$ is defined as:
\[ \textrm{CRPS}(p, \mathbf{x}) = \sum_{i=1}^H \mathbb{E}_{\mathbf{y} \sim p} \left[   |\mathbf{x}_i - \mathbf{y}_i| \right] - \frac{1}{2} \mathbb{E}_{\mathbf{y}, \mathbf{z} \sim p} \left[ |\mathbf{y}_i - \mathbf{z}_i| \right] \]
To evaluate the correlation structure of the joint predictive, we use the \emph{variogram score} (VS):
\[ \textrm{VS}(p, \mathbf{x}) = \sum_{i=1}^H \sum_{j=1}^H \left( |\mathbf{x}_{i} - \mathbf{x}_j|^{0.5} - \mathbb{E}_{\mathbf{y} \sim p} |\mathbf{y}_i - \mathbf{y}_j|^{0.5} \right)^2 \]
Realistic multi-step predictions must achieve low scores on both CRPS and VS \citep{variogram}.

\begin{figure}[t]
    \centering
    \includegraphics[width=1\linewidth]{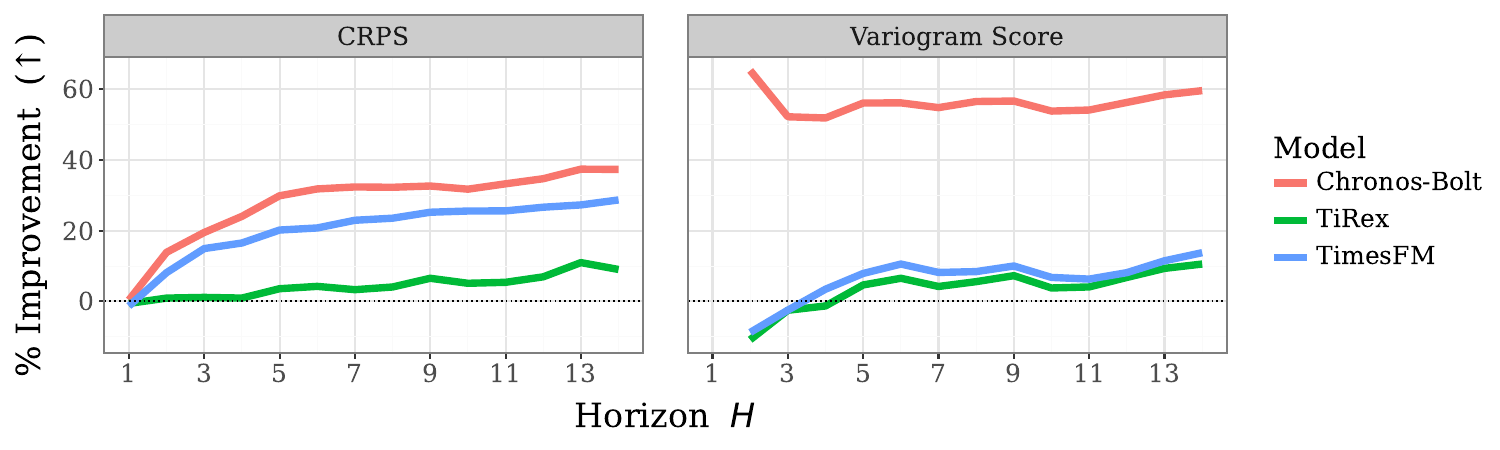}
    \caption{Median percent improvement from copula-based method by horizon for M4 Daily dataset. Switching from autoregressive sampling to our copula-based approach yields high-quality sample paths with realistic correlation structures, and significant improvements in CRPS for longer forecast horizons. Here, we plot the CRPS term at each horizon separately, rather than the cumulative sum.}
    \label{snowball}
    \vspace{-0.3cm}
\end{figure}

\section{Efficiently generating sample paths with multi-step forecasting models}

We generate 10 sample paths with each approach for time series in the M1, M3, M4, and Tourism collections from the Monash repository \citep{monash}. These datasets include diverse domains, granularities, and forecast horizons. See Table \ref{table:datasets} for more details on these datasets.

To show that our copula-based approach can be effectively applied to a variety of backbone models, we evaluate performance using several TSFMs with varying architectures: the decoder-only transformer TimesFM \citep{timesfm}, the encoder-decoder transformer Chronos-Bolt Tiny \citep{chronos_bolt}, and the xLSTM-based TiRex \citep{tirex}. While M4 data was included in TimesFM's training corpus, this does not affect our conclusions as we primarily contrast copula-based and autoregressive sampling for a given TSFM backbone.

In Figure \ref{snowball}, we show the percent improvement in quality for the copula-based sample baths compared to those generated via autoregressive sampling for the M4 Daily dataset; we show similar patterns hold across other datasets in Appendix \ref{full_results}. The copula-based methods produce sample paths with comparable or improved correlation structures compared to autoregressive sampling, as measured by VS. This finding confirms that our simple copula parameterization effectively captures the correlation structures prevalent in time series.

More surprisingly, the copula-based methods also improve the quality of the marginal predictive distributions, as measured by CRPS. We attribute this to the \emph{snowballing error} phenomenon: biases appearing early in generation can amplify biases later on in the forecast horizon. This snowballing error phenomenon is one main argument in favor of multi-token prediction objectives in language modeling \citep{pitfalls}.

\begin{figure}[t!]
    \centering
    \includegraphics[width=1\linewidth]{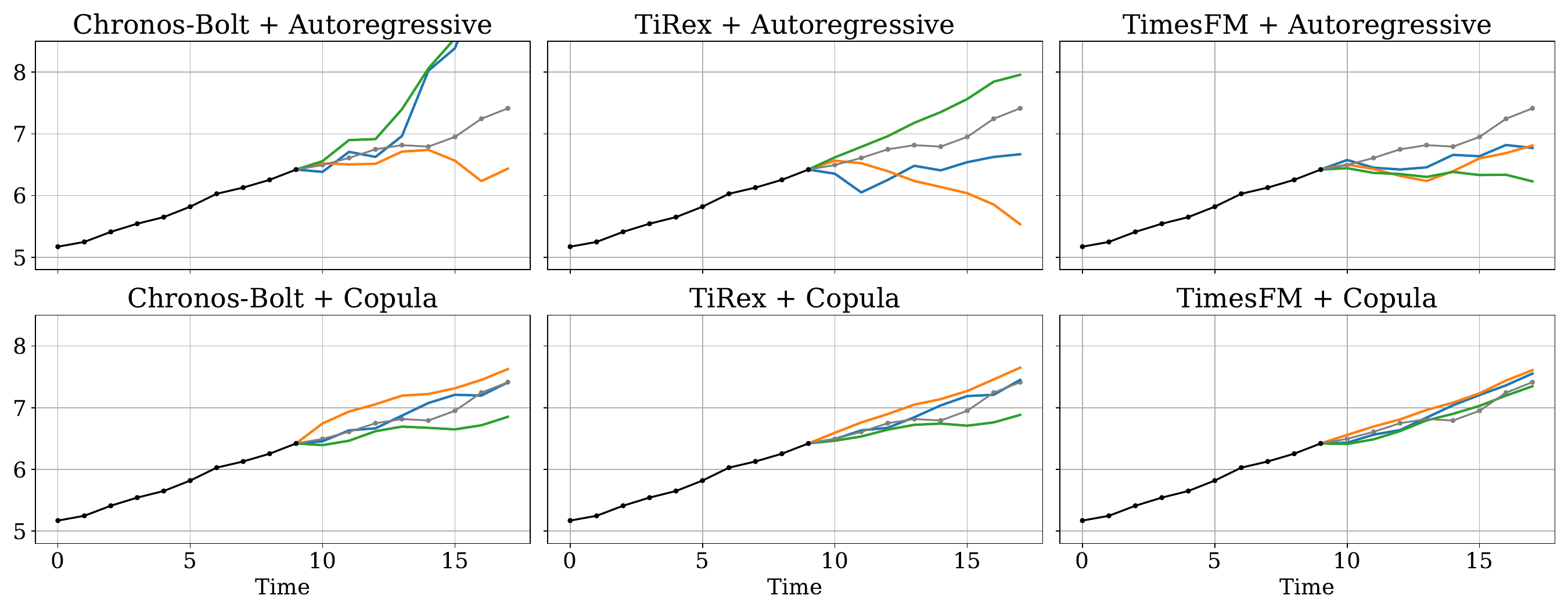}
    \caption{Copula-based sampling mitigates snowballing errors in autoregressive sampling. For each method, we plot 3 random sample paths. The true future trajectory of the time series is shown in gray.}
    \label{fig:snowball}
\end{figure}

In Figure \ref{fig:snowball}, we contrast autoregressive and copula-based sample paths for an input time series with an approximately linear increasing trend. The autoregressively-generated sample paths frequently deviate from the approximately linear trend of the input time series, such as failing to continue the monotonically increasing pattern, or introducing sudden exponential growth, a result of snowballing errors. In contrast, the copula-generated sample paths realistically approximate the true trajectory.

Lastly, recall that generating sample paths via autoregressive sampling takes $\mathcal{O} (N \cdot H)$ time, whereas the copula-based approach allows us to generate arbitrarily-many sample paths in $\mathcal{O}(1)$ time. Unsurprisingly, we therefore observe that the speed of the copula-based methods is orders of magnitude faster than the autoregressive sampling approach. For example, for the M4 Daily dataset, copula-based sampling yielded speedups of 3.7x, 72.4x, and 100.4x for Chronos-Bolt, TiRex, and TimesFM respectively. Crucially, this speedup allows practitioners to adopt our approach for multi-step sample path generation in large-scale production settings. For example, generating correlated sample paths for many time series with TabPFN-TS \citep{tabpfn_ts} would be prohibitively slow using autoregressive sampling, but becomes practical with our method.

\section{Conclusion}

We present a copula-based technique to generate high-quality sample paths from multi-step TSFMs in one forward pass. We verify that our method produces realistic correlation structures and can even improve the accuracy of marginal predictive distributions compared to autoregressive sampling, at a fraction of the computational cost. This technique has potential to greatly accelerate the impact of TSFMs by extending their powerful capabilities to settings requiring joint predictive distributions.


\newpage
{\small
\bibliography{references}
}


\newpage
\appendix
\renewcommand{\thefigure}{A\arabic{figure}}
\setcounter{figure}{0}
\renewcommand{\thetable}{A\arabic{table}}
\setcounter{table}{0}

\section{Full results for all datasets}
\label{full_results}

\begin{figure}[h!]
    \centering
    \includegraphics[width=1\linewidth]{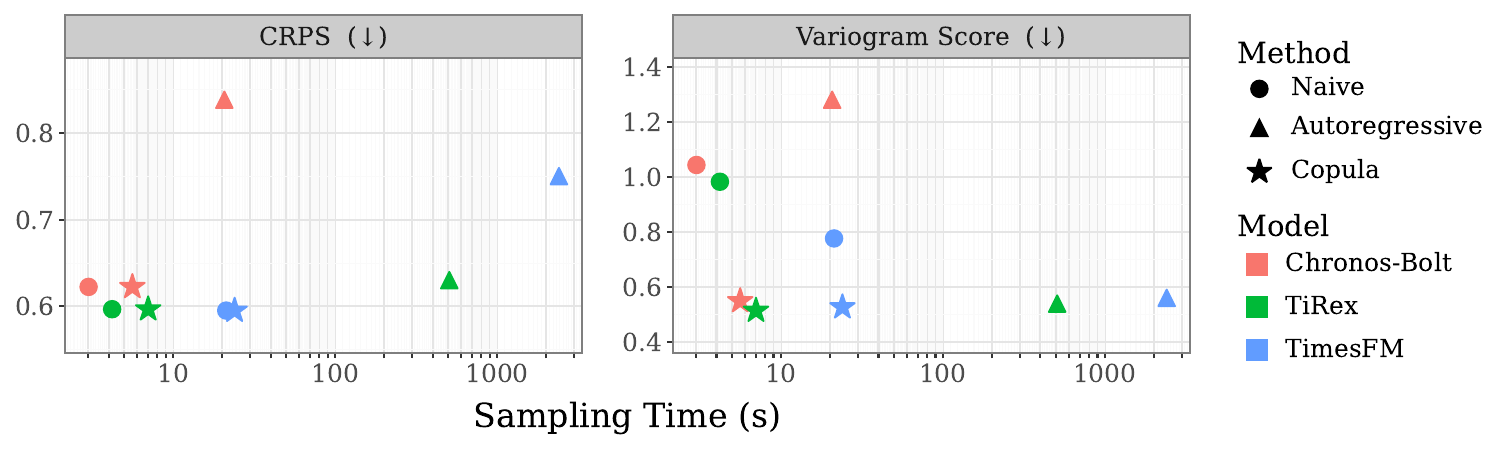}
    \caption{Sample path quality for M4 Daily dataset. Switching from autoregressive sampling to our copula-based approach yields improved sample paths with realistic correlation structures, at a fraction of the time. Points show the median performance across series after dividing by the corresponding score of the SeasonalNaive baseline. The $x$-axis indicates the time taken to generate 10 sample paths per time series on an A100 GPU.}
    \label{fig:m4_daily}
\end{figure}

\begin{figure}[t!]
    \centering
    \includegraphics[width=1\linewidth]{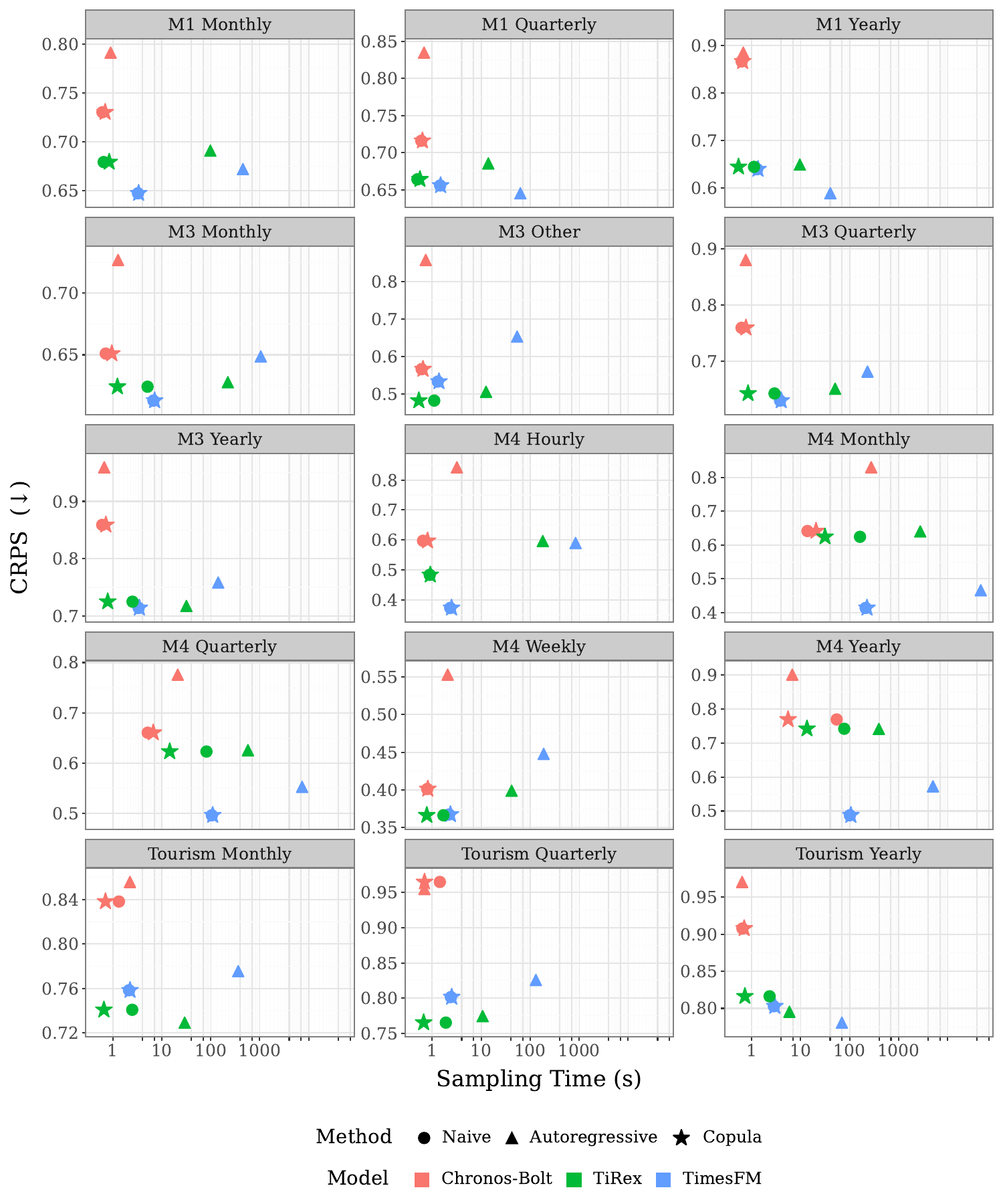}
    \caption{Median CRPS results for other datasets. The copula approach yields sample paths with comparable or higher quality than autoregressive sampling but is significantly faster.}
    \label{crps}
\end{figure}

\begin{figure}[t!]
    \centering
    \includegraphics[width=1\linewidth]{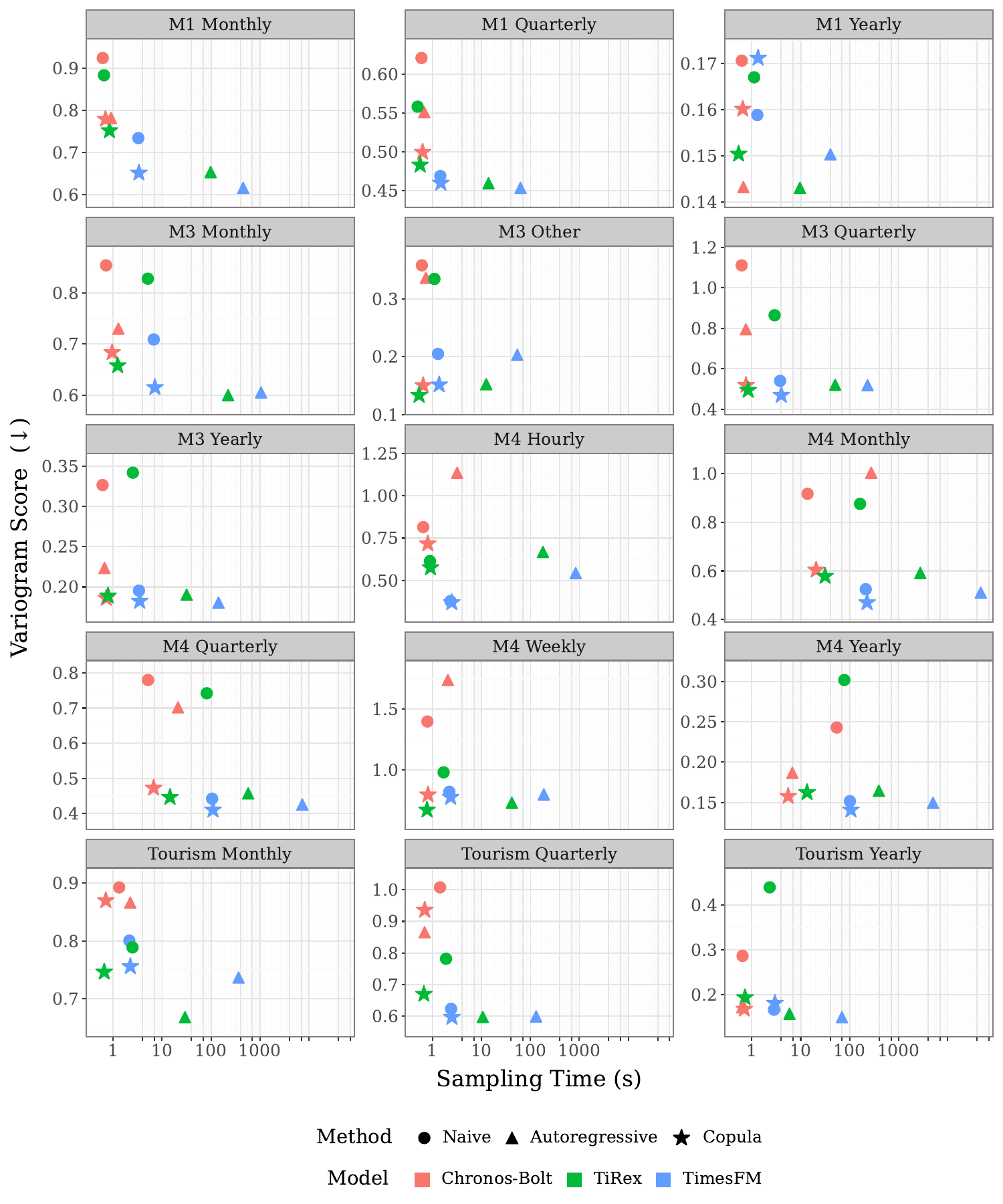}
    \caption{Median variogram score results for other datasets. The copula approach yields sample paths with comparable quality to autoregressive sampling but is significantly faster.}
    \label{vs}
\end{figure}

In Figure \ref{fig:m4_daily}, we contrast the quality of copula-based sample paths to those generated via autoregressive sampling for the M4 Daily dataset, while we show full results for other datasets in Figures \ref{crps} and \ref{vs}. The copula-based methods generally produce sample paths with comparable or improved correlation structures compared to autoregressive sampling, as measured by VS. The copula-based methods also frequently improve upon the marginal predictive distributions from autoregressive sampling, as measured by CRPS. The properties of these datasets are summarized in Table~\ref{table:datasets}.

\vspace{10mm}

\vspace{10mm}

\begin{table}[h!]
\caption{Summary of dataset properties.}
\label{table:datasets}
\centering
\footnotesize
\begin{tabular}{llccccc}
\toprule
& Frequency & Seasonality & Horizon & Series & Min Length & Max Length \\ \midrule
\multirow{3}{*}{M1} 
& Monthly   & 12 & 18 & 617   & 48  & 150 \\
& Quarterly & 4  & 8  & 203   & 18  & 114 \\
& Yearly    & 1  & 6  & 181   & 15  & 58  \\ \midrule
\multirow{4}{*}{M3} 
& Other     & 1  & 8  & 174   & 71  & 104 \\
& Monthly   & 12 & 18 & 1428  & 66  & 144 \\
& Quarterly & 4  & 8  & 756   & 24  & 72  \\
& Yearly    & 1  & 6  & 645   & 20  & 47  \\ \midrule
\multirow{6}{*}{M4} 
& Hourly    & 24 & 48 & 414    & 748   & 1008 \\
& Daily     & 7  & 14 & 4227   & 107   & 9933 \\
& Weekly    & 52  & 13 & 359    & 93    & 2610 \\
& Monthly   & 12 & 18 & 48000  & 60    & 2812 \\
& Quarterly & 4  & 8  & 24000  & 24    & 874  \\
& Yearly    & 1  & 6  & 23000  & 19    & 841  \\ \midrule
\multirow{3}{*}{Tourism} 
& Monthly   & 12 & 24 & 366   & 91  & 333 \\
& Quarterly & 4  & 8  & 427   & 30  & 130 \\
& Yearly    & 1  & 4  & 518   & 11  & 47  \\ \bottomrule
\end{tabular}
\end{table}

\clearpage
\section{Pre-trained copula module}
\label{app:pcm}

Above, we parameterize the Gaussian copula with a single parameter $\rho$, and obtain $\rho$ by computing the autocorrelation coefficient for each time series (``Auto-$\rho$''). Here, we explore the idea of a \emph{pre-trained copula module}, a lightweight neural network trained to predict the Gaussian copula parameters for a given time series. As the backbone of the neural network, we consider:
\begin{enumerate}
    \item An MLP with hidden sizes 64 and 32. The MLP takes as input the last 30 time steps.
    \item A Temporal Convolution Network (TCN) \citep{tcn} with hidden state size of 16, kernel size of 2, dilations of [1, 2, 4, 8]. The final hidden state is passed through an MLP with hidden size of 16 to predict the copula parameters. The TCN takes as input the last 30 time steps.
    \item A Gated Recurrent Unit (GRU) \citep{gru}, with hidden state size of 16. The final hidden state is passed through an MLP with hidden size of 16 to predict the copula parameters. The GRU takes at most the last 128 time steps as input and can handle shorter sequences as well.
\end{enumerate}
For each backbone, we consider two possible parameterizations for the copula covariance matrix $\Sigma$:
\begin{enumerate}
    \item Predicting $\rho$ and setting $\Sigma_{ij} = \rho^{|i-j|}$ as above.
    \item Predicting $\rho$ and $\beta$, and setting $\Sigma_{ij} = (1-\beta) \rho^{|i-j|} + \beta \delta_{ij}$. This parameterization is indicated with \verb|XXX-2|.
\end{enumerate}
We train the backbones for 10 epochs on M1 Monthly, M4 Hourly, M4 Daily, M4 Weekly, M4 Monthly, and Tourism Monthly. For each series, we generate 20 sample paths up to a horizon of $H=8$ and minimize the variogram score using Adam with a learning rate of 0.001. We scale each input time series by dividing by its maximum value. During training, we use the marginal predictive distributions of Chronos-Bolt Tiny.

\begin{figure}[h!]
    \centering
    \includegraphics[width=1\linewidth]{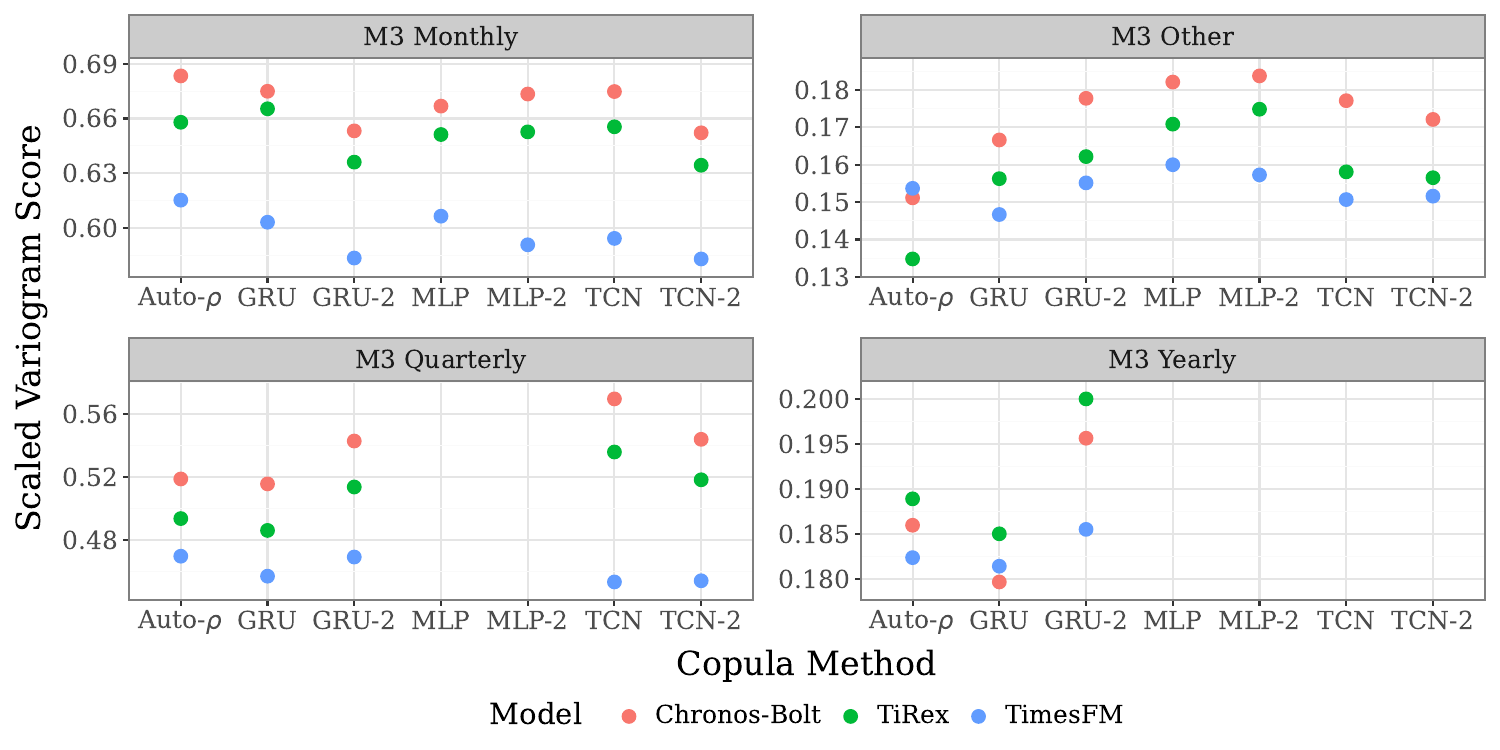}
    \caption{Median variogram score results for M3 datasets with various pre-trained copula modules, normalized by the SeasonalNaive score. Since the MLP and TCN backbones have constraints for the input sequence lengths, their scores are not available for some datasets.}
    \label{pcm}
\end{figure}

In Figure \ref{pcm}, we compare the performance of each pre-trained copula module in generating accurate correlation structures on the M3 datasets. 

The pre-trained copula modules sometimes yield improved sample paths compared to the ``Auto-$\rho$'' approach presented above. In particular, the GRU backbone seems most promising, improving performance on three of the four evaluation datasets, and is also capable of handling inputs with varying lengths. Interestingly, despite being trained solely on the marginal predictions of Chronos-Bolt Tiny, the pre-trained copula modules often yield similar improvements in performance when applied to the marginal predictive distributions of other models. This observation reinforces the promise of applying a pre-trained copula module to cheaply augment the predictions of any multi-step TSFM and produce correlated sample paths. 

Lastly, while switching to the $\rho, \beta$ parameterization provides more flexibility, in our experiments this does not translate into improved performance in practice. We suspect that more extensive pre-training of the modules, potentially including synthetic training data and longer forecast horizons, could address this shortcoming. Moreover, future research could explore more expressive parameterizations for Gaussian copulas, or even other types of copula structures.

\vspace{6mm}
\section{Fitting empirical CDF to quantile predictions}
\label{empirical_cdf}

\begin{figure}[h!]
    \centering
    \includegraphics[width=0.6\linewidth]{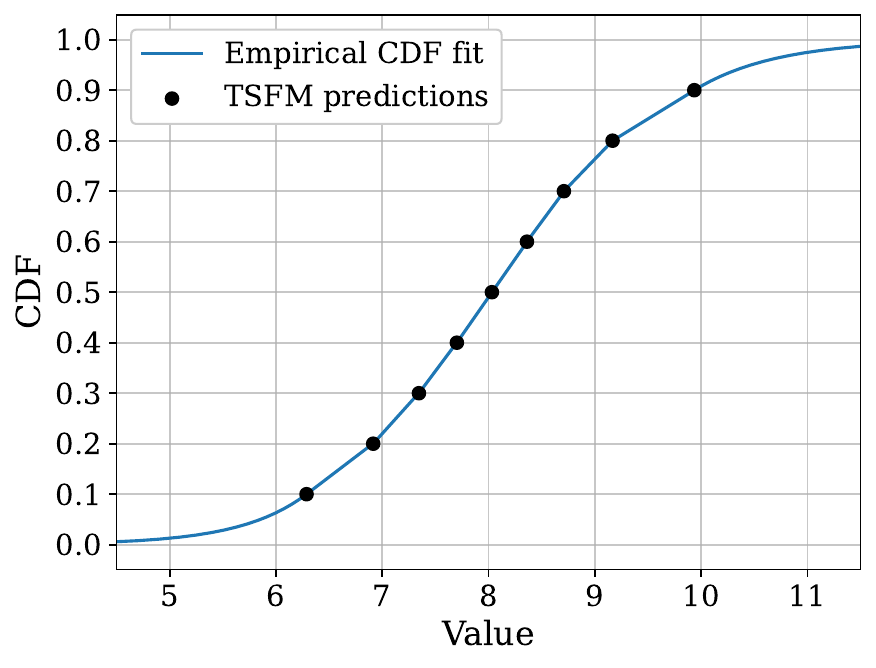}
    \caption{Example of empirical CDF fit to TSFM quantile predictions using IQF.}
    \label{fig:iqf}
\end{figure}

The TSFMs we consider in this paper produce marginal predictions in the form of quantile predictions for $Q = \{0.1, 0.2, \ldots, 0.9\}$. To convert these predictions to full marginal distributions, we adopt the incremental quantile function (IQF) approach from \cite{iqf}. In Figure \ref{fig:iqf}, we show an example of an empirical CDF fit to quantile predictions from a TSFM.

\vspace{6mm}

\end{document}